\documentclass[conference]{IEEEtran}
\ifCLASSOPTIONcompsoc
  \usepackage[nocompress]{cite}
\else
  \usepackage{cite}
\fi
\ifCLASSINFOpdf
\else
\fi

\usepackage{svg}
\usepackage{color,soul}
\usepackage{hyperref}
\usepackage{amsmath}
\usepackage{float}
\usepackage{multirow}

\usepackage{color}
\usepackage{soul}
\usepackage{stfloats}

\begin{document}
%
% paper title
% Titles are generally capitalized except for words such as a, an, and, as,
% at, but, by, for, in, nor, of, on, or, the, to and up, which are usually
% not capitalized unless they are the first or last word of the title.
% Linebreaks \\ can be used within to get better formatting as desired.
% Do not put math or special symbols in the title.
\title{ProGNNosis: A Data-driven Model to Predict GNN Computation Time Using Graph Metrics}
%
%
% author names and IEEE memberships
% note positions of commas and nonbreaking spaces ( ~ ) LaTeX will not break
% a structure at a ~ so this keeps an author's name from being broken across
% two lines.
% use \thanks{} to gain access to the first footnote area
% a separate \thanks must be used for each paragraph as LaTeX2e's \thanks
% was not built to handle multiple paragraphs
%
%
%\IEEEcompsocitemizethanks is a special \thanks that produces the bulleted
% lists the Computer Society journals use for "first footnote" author
% affiliations. Use \IEEEcompsocthanksitem which works much like \item
% for each affiliation group. When not in compsoc mode,
% \IEEEcompsocitemizethanks becomes like \thanks and
% \IEEEcompsocthanksitem becomes a line break with idention. This
% facilitates dual compilation, although admittedly the differences in the
% desired content of \author between the different types of papers makes a
% one-size-fits-all approach a daunting prospect. For instance, compsoc 
% journal papers have the author affiliations above the "Manuscript
% received ..."  text while in non-compsoc journals this is reversed. Sigh.

\author{\IEEEauthorblockN{Axel Wassington}
\IEEEauthorblockA{\textit{Barcelona Neural Networking Center (BNN)} \\
\textit{Universitat Politècnica de Catalunya (UPC)}\\
Barcelona, Spain \\
axel.tomas.wassington@upc.edu}
\and
\IEEEauthorblockN{Sergi Abadal}
\IEEEauthorblockA{\textit{Barcelona Neural Networking Center (BNN)} \\
\textit{Universitat Politècnica de Catalunya (UPC)}\\
Barcelona, Spain \\
abadal@ac.upc.edu}
}

\markboth{Submitted to IEEE Computer Architecture Letters}%
{Wassington, Abadal: Computation Time Prediciton of Graph Neural Networks via Graph Characterization}
% The only time the second header will appear is for the odd numbered pages
% after the title page when using the twoside option.
% 
% *** Note that you probably will NOT want to include the author's ***
% *** name in the headers of peer review papers.                   ***
% You can use \ifCLASSOPTIONpeerreview for conditional compilation here if
% you desire.

% The publisher's ID mark at the bottom of the page is less important with
% Computer Society journal papers as those publications place the marks
% outside of the main text columns and, therefore, unlike regular IEEE
% journals, the available text space is not reduced by their presence.
% If you want to put a publisher's ID mark on the page you can do it like
% this:
%\IEEEpubid{0000--0000/00\$00.00~\copyright~2015 IEEE}
% or like this to get the Computer Society new two part style.
%\IEEEpubid{\makebox[\columnwidth]{\hfill 0000--0000/00/\$00.00~\copyright~2015 IEEE}%
%\hspace{\columnsep}\makebox[\columnwidth]{Published by the IEEE Computer Society\hfill}}
% Remember, if you use this you must call \IEEEpubidadjcol in the second
% column for its text to clear the IEEEpubid mark (Computer Society journal
% papers don't need this extra clearance.)

% use for special paper notices
%\IEEEspecialpapernotice{(Invited Paper)}

% for Computer Society papers, we must declare the abstract and index terms
% PRIOR to the title within the \IEEEtitleabstractindextext IEEEtran
% command as these need to go into the title area created by \maketitle.
% As a general rule, do not put math, special symbols or citations
% in the abstract or keywords.
\IEEEtitleabstractindextext{%
\begin{abstract}
Graph Neural Networks (GNN) show great promise in problems dealing with graph-structured data. One of the unique points of GNNs is their flexibility to adapt to multiple problems, which not only leads to wide applicability, but also poses important challenges when finding the best model or acceleration technique for a particular problem. An example of such challenges resides in the fact that the accuracy or effectiveness of a GNN model or acceleration technique, respectively, generally depends on the structure of the underlying graph. 
In this paper, in an attempt to address the problem of graph-dependent acceleration, we propose \textsc{ProGNNosis}, a data-driven model that can predict the GNN training time of a given GNN model running over a graph of arbitrary characteristics by inspecting the input graph metrics. Such prediction is made based on a regression that was previously trained offline using a diverse synthetic graph dataset. In practice, our method allows making informed decisions on which design to use for a specific problem. In the paper, the methodology to build \textsc{ProGNNosis} is defined and applied for a specific use case, where it helps to decide which graph representation is better. Our results show that \textsc{ProGNNosis} helps achieve an average speedup of 1.22$\times$ over randomly selecting a graph representation in multiple widely used GNN models such as GCN, GIN, GAT, or GraphSAGE.
\end{abstract}

% Note that keywords are not normally used for peer review papers.
}

% make the title area
\maketitle

% To allow for easy dual compilation without having to reenter the
% abstract/keywords data, the \IEEEtitleabstractindextext text will
% not be used in maketitle, but will appear (i.e., to be "transported")
% here as \IEEEdisplaynontitleabstractindextext when compsoc mode
% is not selected <OR> if conference mode is selected - because compsoc
% conference papers position the abstract like regular (non-compsoc)
% papers do!
\IEEEdisplaynontitleabstractindextext
% \IEEEdisplaynontitleabstractindextext has no effect when using
% compsoc under a non-conference mode.

% For peer review papers, you can put extra information on the cover
% page as needed:
% \ifCLASSOPTIONpeerreview
% \begin{center} \bfseries EDICS Category: 3-BBND \end{center}
% \fi
%
% For peerreview papers, this IEEEtran command inserts a page break and
% creates the second title. It will be ignored for other modes.
\IEEEpeerreviewmaketitle

\ifCLASSOPTIONcompsoc
\IEEEraisesectionheading{\section{Introduction}\label{sec:introduction}}
\else
\section{Introduction}
\label{sec:introduction}
\fi
% Computer Society journal (but not conference!) papers do something unusual
% with the very first section heading (almost always called "Introduction").
% They place it ABOVE the main text! IEEEtran.cls does not automatically do
% this for you, but you can achieve this effect with the provided
% \IEEEraisesectionheading{} command. Note the need to keep any \label that
% is to refer to the section immediately after \section in the above as
% \IEEEraisesectionheading puts \section within a raised box.

% The very first letter is a 2 line initial drop letter followed
% by the rest of the first word in caps (small caps for compsoc).
% 
% form to use if the first word consists of a single letter:
% \IEEEPARstart{A}{demo} file is ....
% 
% form to use if you need the single drop letter followed by
% normal text (unknown if ever used by the IEEE):
% \IEEEPARstart{A}{}demo file is ....
% 
% Some journals put the first two words in caps:
% \IEEEPARstart{T}{his demo} file is ....
% 
% Here we have the typical use of a "T" for an initial drop letter
% and "HIS" in caps to complete the first word.
%% Generic intro
\IEEEPARstart{G}{raph} Neural Networks (GNNs) have recently attracted enormous interest in the machine learning community due to their ability to infer from graph-structured data, which other types of neural networks cannot handle efficiently \cite{wu2020comprehensive}. This has a transformative impact on areas such as recommendation systems \cite{10.1145/3535101}, natural language processing \cite{nlp}, computer vision \cite{shi2020point}, particle physics \cite{ju2020graph}, or computer networks \cite{rusek2020routenet, ferriol2022routenet} as the explosion of recent works can attest. 

%% Generic challenge intro
One of the strengths of GNNs, which make them applicable to a wide variety of problems in multiple domains, is their unique structure and flexibility \cite{zhou2020graph}. Indeed, GNNs can be understood as a family of algorithms allowing to infer features relative to single vertices, edges, or the entire aggregated graphs. However, also due to their wide applicability, not only creating a single GNN model that fits all the scenarios is rendered very difficult, but even the selection of the most accurate GNN model in a specific scenario already becomes a complex and multidimensional problem. 

%%The use of GNNs for different domains of knowledge, and the different characteristics of the problems of each domain, has made a great deal of different models and accelerators to handle the needs of the different domains \cite{zhou2020graph}. \hl{Owing to their unique structure, GNNs can be understood as a family of algorithms. Also, they present unique challenges in their efficient processing due to the alternate execution of dense and very sparse operations, as well as their dependence on the input graph} \cite{abadal2021computing}. \hl{As a result, a myriad of works have studied GNNs from a computational workload perspective and attempted to extract the architectural implications of their uniquenesses, aiming to improve their support in CPUs, GPUs, and more advanced accelerators.} \cite{Yan2020, Zhang2020b, abadal2021computing,liu2022survey}.

%% Challenge more specific: computation.
In exchange for such flexibility, GNNs also present unique challenges in their efficient processing due to the variety of GNN variants to support, the inherently alternate execution of dense and very sparse operations, or their dependence on the input graph \cite{abadal2021computing}. As a result, recent works have studied GNNs from a computational workload perspective and attempted to extract the architectural implications of their uniquenesses, aiming to improve their support in CPUs, GPUs, and hardware accelerators \cite{yan2020characterizing, Zhang2020b, garg2022understanding, abadal2021computing, liu2022survey}. Yet still, solutions that generalize well over all GNN variants and application domains are missing.

Multiple attempts to either maximize the accuracy of a GNN or minimize its processing time have been made. For example, the impact of some design decisions (e.g. the GNN model, the hidden vector size, or the loss function) on the accuracy given a GNN solution has been studied using techniques such as a guided search \cite{zhou2019auto, lai2020policy, wang2021automated, you2020design}. A more comprehensive approach is to perform a joint search on both the GNN model and the acceleration technique to navigate the accuracy-processing time tradeoffs of the solution space \cite{zhang2021g}. However, these works use search methods and try different combinations to then keep the most performant ones. This approach has the problem of spending the time to try different strategies in an \emph{ad hoc} manner. Another rather new approach is to use a synthetic dataset to see the relationship between the parameters of the graph generator and the accuracy of the different GNN models \cite{palowitch2022graphworld}, this allows to explore the relationship between a limited set of characteristics of the graph with the accuracy of the different models and can be used as a benchmarking dataset.

% State of the art (II)
For the analysis of the computation time of GNNs, works are generally infrequent and cover a small design space. One example is the study made in \cite{yan2020characterizing}, where the impact of the input and output feature vector size is studied for a small selection of GNNs models and datasets. In \cite{guirado2021characterizing}, the authors model the performance and resource efficiency of two hardware accelerators for GNN as functions of several hardware parameters and GNN models but neglecting the irregular connectivity of real graphs. A more comprehensive analysis of the most popular frameworks and GNN models is done in \cite{9408211}, which uses the Open Graph Benchmark (OGB) \cite{hu2020ogb} dataset to make a comparison among GNN acceleration techniques. 

%One way to assess the performance of GNN processing, also the one used in this work, is to measure the time used to train a single epoch. 
%This metric is widely used to compare the result of new GNN acceleration techniques. 

% Gap in the literature 
Although the OGB suite used in multiple GNN comparison works contains a representative selection of datasets for real-world problems, the span of the suite is limited with most graphs having similar structural characteristics and connectivity, hence failing to cover all possible future application domains. 
%Hence, a wide variety of problems from different fields may not be well represented on this kind of small dataset. 
This is an issue due to the dependence of GNNs on the input graph: the accuracy and computation time of a GNN, both in training and inference, generally depends on the \emph{aggregation} of the features from each node to its neighbors, which is generally extremely sparse and irregular process.
%In essence, this depends on the graph topology as it determines the size of such operation and its sparsity which, in turn, can widely vary across the graph. 
In essence, works largely ignore the characteristics of the input graph as they often consider sparsity as the only aspect affecting performance \cite{Tian2020,wang2021tc}, and only evaluate their work on OGB or a smaller selection of graphs.

\begin{figure*}
    \centering
    \includegraphics[width= 0.95\textwidth]{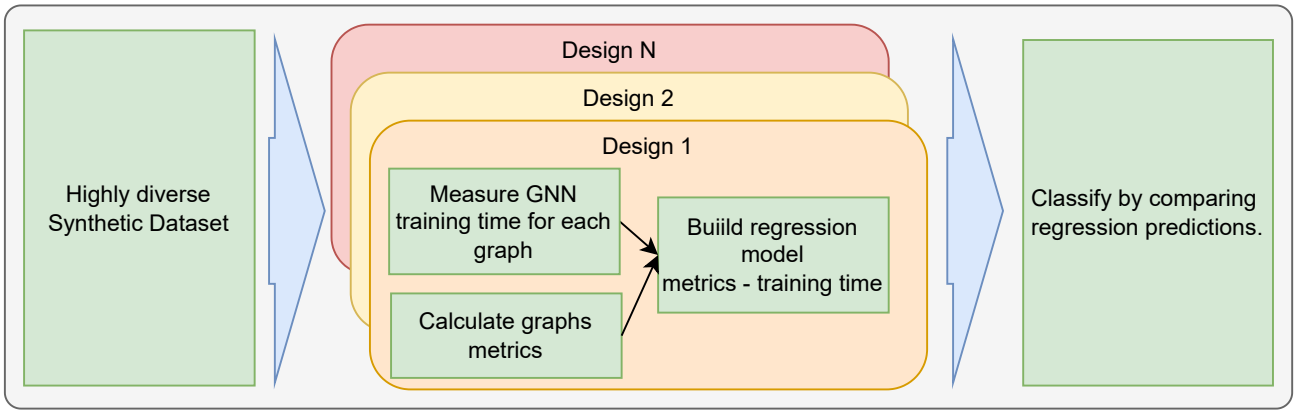}
    \caption{High-level description of \textsc{ProGNNosis} as the main contribution of this paper.}
    \label{fig:abstract}
\end{figure*}

This paper aims to address these issues by proposing \textsc{ProGNNosis} (Fig. \ref{fig:abstract}), a framework to build data-driven models able to predict the computation time of a specific GNN model executed over a given computation platform, but for a graph of arbitrary characteristics. When repeated over multiple models or computing platforms, possibly with the help of prototyping tools \cite{pujol2021ignnition}, \textsc{ProGNNosis} allows assessing which GNN design will perform better for any type of input graph, without having to perform a search as in prior approaches. 

To these ends, \textsc{ProGNNosis} generates a comprehensive and balanced synthetic graph dataset and makes an offline analysis of the impact of the different graph metrics (such as degree distribution and clustering coefficient) on the GNN execution time. Building on this analysis, we demonstrate that a model can be generated that predicts the execution time of the GNN with high accuracy. Then, we establish a simple but representative use case (i.e. the graph representation in memory) and show that we can use \textsc{ProGNNosis} to make informed design decisions about GNN acceleration for any type of graph, achieving a mean speedup of 1.24$\times$ for the training set and 1.07$\times$ for the testing set with respect to making a random choice in a GCN. We finally repeat the experiment with different GNN models (GIN, GAT, GraphSAGE) to show that the proposed methodology is potentially generalizable, obtaining a mean speedup of 1.26$\times$ for the training set and 1.18$\times$ for the testing set.

%Then we demonstrate through a specific case that the model performs well, by first showing it can predict the execution time with XXX accuracy, and then demonstrating that this prediction is enough to decide between two designs. We then repeat the experiment with different GNN models to show that the result is generalizable and analyze the impact of the decision variable on the different models.

The remainder of the paper is organized as follows. The notation and other preliminary considerations are described in Section \ref{sec:prelim}. Our main contribution, a methodology for the data-driven modeling of the execution time of GNNs, is presented in Section \ref{sec:methodology}. The methodology is then illustrated for different use cases in Section \ref{sec:results}. Finally, the paper is concluded in Section \ref{sec:conc}.

% You must have at least 2 lines in the paragraph with the drop letter
% (should never be an issue)

\section{Preliminaries}
\label{sec:prelim}
A graph is an ordered pair $G=(V, E)$, where $V$ is a set of vertices (or nodes) and $E=\{\{u,v\}: u,v \in V\}$ a set of edges that are connections between pairs of nodes. Stemming from this definition, next we describe the graph metrics, GNN models, and regression and classification considerations used in this work.

\subsection{Graph Metrics}
Graphs are complex structures that can be measured from multiple angles. Depending on the objective of the measurement, different characteristics of the graph can take a central role or have no impact at all. Also, some characterization metrics are correlated and others are not. 
%So, in general, graph measurement is a complex world. 
Another important aspect is the time it takes to calculate the metrics: some metrics can be assessed via a quick inspection of the graph, and others need a great number of calculations, which in some cases can be reduced by obtaining approximate estimates.

In this study, we will use undirected, unweighted, and connected graphs for simplicity. The definitions given in this section will only consider this kind of graph, but the results of the study can be generalized to other kinds of graphs.

\vspace{0.2cm}
\noindent
\textbf{Degree:} The degree of a node is the number of vertices incident to that node or, in other words, the number of connections a node has. Hence, the degree $k_v$ of a node $v$ is generally given by the size of its neighborhood,
\begin{equation}
    k_v = |N(v)|,
\end{equation}
%normally referenced as $k_v$ and can be defined as the size of the neighborhood of $v$, $k_v = |N(v)|$.
where the neighborhood of a node can be defined as $N(v) = \{u: \{u,v\} \in V\}.$ 
%The degree sequence is the sequence of the degrees of all the nodes in the graph. 
Then, the degree distribution can be defined as the fraction of nodes with a given degree. Different characterizations can be extracted from the degree distribution, but some of the most useful ones are the maximum degree, the minimum degree, and the mean degree of a graph. The calculation of the degree distribution can be done in linear order concerning the number of edges in most of the graph representations.

\vspace{0.2cm}
\noindent
\textbf{Density:} The density $D$ of a graph is the ratio between the edges that are present in the graph and the maximum amount of edges that the graph may have, given its number of nodes. Hence, the density of an undirected graph can be defined as:
\begin{equation}
    D(G) = \frac{2\left| E \right|}{\left| V \right|(\left| V \right| -1)}.
\end{equation}

The density can be calculated in linear order in most of the graph representations, and most of the time the number of nodes and edges is already calculated and stored with the graph.

\vspace{0.2cm}
\noindent
\textbf{Clustering coefficient}
The clustering coefficient $C(v)$ of a node $v$ can be defined as:
\begin{equation}
    C(v) = \frac{|\{\{u,w\}: (\{u,w\} \in E \land u,w \in N(v)\}|}{k_v (k_v - 1)}
\end{equation}

This indicates how close is the neighborhood of that node to generating a complete subgraph (or clique). The mean clustering coefficient is, as its name indicates, the mean of the clustering coefficient of all the nodes in a graph. 
%It is also called transitivity because of its relation to the transitivity property of a predicate. 
The computational complexity of calculating the clustering coefficient is $O(n^3)$.
%and, for large graphs, this can be too much time for the use we will give to it. 
Because of this, we use an approximation to calculate the clustering coefficient that is based on using trials instead of using all the nodes to calculate the coefficient, based on the ideas proposed in \cite{schank2005approximating}.

\subsection{Regression and Classification}
In this work, we use regression and/or classification to extract the relationship between the computation time of a GNN model and the characteristics of an input graph, defined by some of the metrics described above. 

On the one hand, regression is the statistical process used to find the relationship between a dependent variable and one or more independent variables (or features). The linear regression finds a linear combination of the independent variables that reduce the sum of squared differences with the dependent variable. To measure the performance of regression we will use three metrics:
\begin{itemize}
    \item The \textbf{coefficient of determination ($R^2$)} is the sum of squared residuals and is 1 if all predictions are correct (best case), is 0 for the baseline model (using always the mean as prediction), and maybe negative if the prediction is worst than the baseline model.
    \item The \textbf{mean squared error (MSE)} is the mean of the square of the differences between the predicted and the real values. It is 0 if all predictions are correct and its value increases with the errors. Since it grows with the square of the error, it is a good indicator of outliers.
    \item The \textbf{mean absolute percentage error (MAPE)} is the sum of the errors divided by the real value. MAPE is good to find if the error does grow disproportionate to the real value.
\end{itemize}

On the other hand, classification is also a statistical process, but in this case, the dependent variable is discrete and each value it can take is called a class. One popular option for classification is the use of Support Vector Machine (SVM), which defines a hyperplane that separates the data into categories. This hyperplane is such that it maximizes the margin between the clases. To measure the performance of classification we will use the \textbf{accuracy}, which is simply the number of correctly classified samples divided by the total number of samples.

\subsection{GNN Fundamentals}
\label{sec:preGNN}
%A Graph Neural Network is a Machine learning model that can adapt to graph structures. The main applications of GNNs are node, edge, and graph classification. 
Although GNNs are a wide family of algorithms, in this work we focus on the problem of node classification. The main idea is to use semi-supervised learning, where only some nodes are labeled with their class and the GNN should be able to generalize the labels to the rest of the nodes.

To do so, each node has a vector of input features $x_v$. With this input vector, the GNN does a series of transformations to obtain $z_v$, the output vector. In multiclass node classification, the output vector is composed of one slot for each class, the highest number the one corresponding to the class assigned to the node. Each node goes through a series of intermediate states $h_v^i$. The final state is $h_v^N$, being $N$ the number of layers of the GNN.

In general, the $i$-th layer of a GNN can formally be described as:
\begin{equation}
    h_v^{i+1} = U^i\left(h_v^i, A^i(\{h_u^i: u \in N(v)\}\right)
\end{equation}
where $h_v^i$ is the hidden layer of node $v$ on generation $i$, $A^i(\cdot)$ is the \textit{aggregate} function,
%that is in charge of merging the features of the neighborhood of the node,
whereas $U^i(\cdot)$ is the \textit{update} function that combines the aggregated result with the previous state of the node.

The training of the model is done through backpropagation to minimize a loss function. In each epoch, the weights update function and, in some models, the aggregate function, are updated after the model is used to predict the labels of the labeled nodes. See \cite{abadal2021computing} for more details on the computations of the training process.

%\hl{Esto no va aqui -- yo lo pondria en sec 3 o 4}
%We will compare in this study two ways of representing the graph, that change the way the GNN is processed. One is the sparse adjacency matrix representation (SPARSE), and the other one is the edge list representation (EDGE\_LIST). The SPARSE representation uses sparse matrix multiplication to compute the aggregation function, and the EDGE\_LIST representation uses \textit{gather} and \textit{scatter} CUDA instructions to compute the aggregation function.

\subsection{Sample GNN Models}
We will now present some of the most popular GNN models, which we employ in this work to evaluate the generalization potential of \textsc{ProGNNosis}. The implementation used in this study for all the models is based on Pytorch Geometric (PyG) \cite{fey2019fast}, which is one of the most used frameworks. %, more because of the high level it offers to manage the GNNs than for its performance.

\vspace{0.2cm}
\noindent
\textbf{Graph Convolutional Network (GCN):}
GCN is based on convolutional neural networks, but applied to graphs \cite{kipf2016semi}. It uses a local approximation of the eigenvalues of the adjacency matrix to compute a convolution in the non-euclidean space defined by the graph. The GCN step for a node $v$ can be defined as
\begin{equation}
    h_v^{i+1} = \theta \left( W^i \sum_{u \in N(v) \cup {v}} \frac{h_u^i}{\sqrt{k_u k_v}}  \right),
\end{equation}
where $W$ are the trainable weights and $\theta$ is a function that introduces a non-linearity, e.g. \emph{ReLu}. Note that the aggregated features are normalized to the degree of the connected vertices, e.g. $k_v$.

\vspace{0.2cm}
\noindent
\textbf{Graph Isomorphism Network (GIN):} GIN is a simple model built to demonstrate that even simple models may be powerful on GNN \cite{xu2018powerful}, and aims to classify graphs based on their similarity. The GIN step can be defined as
\begin{equation}
    h_v^{i+1} = MLP^i \left( (1 + \epsilon) h_v + \sum_{u \in N(v)} h_u^i \right),
\end{equation}
where $MLP^i$ is a multi-layer perceptron, and $\epsilon$ is a parameter of the model that indicates the importance of the nodes own value to the ones of its neighborhood.

\vspace{0.2cm}
\noindent
\textbf{Graph Attention Networks (GAT):}
GAT is based on the concept of attention, where the edges have a learnable weight that changes over the generations depending on the feature vectors of the nodes \cite{velivckovic2017graph}. The GAT step can be defined as
\begin{equation}
     h_v^{i+1} = \theta \left( \sum_{u \in N(v) \cup {v}} a_{u,v} W^i \times h_u^i \right),
\end{equation}
where $a_{u,v}$ is the attention coeficient for nodes $u$ and $v$. The attention coefficient can be calculated as
\begin{equation}
    a_{u,v} = softmax_{N(v)}\left(a(W^i \times h_u,W^i \times h_v)\right),
\end{equation}
where $a$ is the attention function, softmax is the normalization between neighbors, and $W$ are the trainable weights.

\vspace{0.2cm}
\noindent
\textbf{GraphSAGE (SAGE):} GraphSAGE (SAmple and aggreGatE) owes its name to the fact that it samples the neighbors of a node for the aggregation stage \cite{hamilton2017inductive}. SAGE can be used with different aggregations, we will present here the one we used that is the \textit{sum} aggregation, which take the form:
\begin{equation}
     h_v^{i+1} =   \theta \left(W^i_1 h_v^i + \sum_{u \in N(v)} W^i_2 h_u^i \right),
\end{equation}
where $W_1$ and $W_2$ are the trainable weights.

\section{ProGNNosis: Approach and Evaluation Methodology}
\label{sec:methodology}

%\hl{In no particular order: Description of the overall methodology, tools used, and how they interrelate. This section should contain details on how data was generated, which benchmarks were used (if any), which libraries, which GNNs, the characteristics of the server where the GNNs were run and the inference time recorded. We need to mention that we do not care about the accuracy of the GNN model, but rather how much time it takes. }

%The main idea behind this methodology is that it is possible
The proposed framework, called \textsc{ProGNNosis}, is able to predict how long will it take for a given GNN to process a certain GNN model in a given computing platform, considering a graph of arbitrary characteristics. With \textsc{ProGNNosis}, we propose to make informed decisions on GNN acceleration or GNN design towards minimizing the training time per generation. %without caring about the accuracy of the GNN inference.

To achieve this objective, we consider a series of steps that are represented in Figure \ref{fig:abstract}:
\begin{enumerate}
    \item The first step consists in creating a synthetic dataset containing a broad set of graphs of different characteristics. This dataset will help us understand the relationship between different graphs and the training time of a GNN. 
    \item Using the dataset, we measure the training time for each of the designs that we want to compare for each of the graphs of the dataset.
    \item In parallel, we calculate a set of characterization metrics over the graphs of the dataset. 
    \item Once we have the data, an analysis can be made to understand the correlations between the metrics and the training time of the different models. In particular, a regression model can be built for each of the designs. 
    \item Finally, the results of the different designs can be compared to classify the graphs into groups that indicate which design performed better. The classification results then are used to corroborate that processing time can be saved by applying this methodology.
\end{enumerate}

Each step of the process can be verified against a testing set, which contains real-world graphs, to see if (i) the synthetic dataset covers the full extent of the testing dataset, (ii) the regression extrapolates to them, and (iii) the classification works correctly with them. The testing set was created using the SparseMatrix collection \cite{Kolodziej2019} and a subset of the OGB benchmarks. Graphs were selected to represent the widest selection of domain and graph characteristics.

The resulting set of models can then be used to optimize the processing of other graphs by predicting the processing time of each design using the pre-trained models and using the one that was predicted to perform better.

\subsection{Synthetic Graph Dataset Generation} 
\label{sec:gen}

\begin{figure}
    \includegraphics[width=\columnwidth]{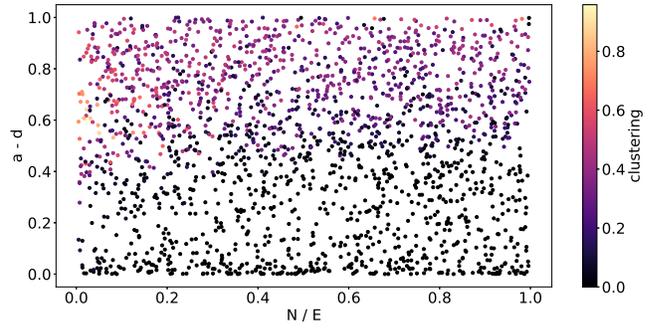}
    \vspace{-0.5cm}
    \caption{Impact of the RMAT parameters on the clustering of the generated graphs. The x-axis corresponds to parameters $N$ and $E$. The way that the dataset was generated makes the distribution of this variable close to uniform ($N$ and $E$ are not the final number of edges and nodes, but the model parameters). The y axis corresponds to the maximum difference between the components of the vector $r$ ($a-d$). The colors indicate the mean clustering coefficient of each graph.}
    \label{fig:param_clustering_original}
\end{figure}

One of the most important aspects of the proposed approach is the creation of a synthetic dataset with diverse characteristics. To create this synthetic dataset, we started by creating a dataset using a naive approach. Then, an optimization problem was stated using this dataset to find a less biased dataset.

To generate the synthetic graphs, the popular RMAT graph generator was used \cite{chakrabarti2004r}. We selected this tool because it can generate graphs with a wide variety of metrics that can be controlled through its parameters, and also because it is quite performant allowing us to generate large datasets in a short time.

The RMAT generator uses six parameters. The first two parameters are the number of nodes $N$ and the number of edges $E$. Then, we have a vector $r = [a, b, c, d]$ of symmetric parameters that define the fitness distribution for the edge attachment. The $r$ vector is a probability vector, and as such the sum of its elements must be 1. The RMAT generator works by dividing the adjacency matrix into four groups recursively and assigning the probability of an edge falling in each of the groups following the $r$ vector.

As already stated in the introduction, one aspect that we found to be especially important in the synthetic dataset used as the training set is that graphs with different characteristics are represented. This means that the dataset should have a wide range and balanced representation of values on the different characterization metrics, hence avoiding selection bias (i.e. bias towards a certain combination of metrics). However, we found that datasets generated in a naive way using RMAT had a selection bias towards graphs with low clustering coefficient, among other characteristics, as can be seen in Figure \ref{fig:param_clustering_original}. 

% TODO: Aggregar referencia a paper de graphlaxy
To overcome this bias, we propose a method where an optimization problem is defined to find an RMAT parameter distribution that generates a less biased dataset \cite{Wassington2022BiasRV}. Also, a tool to train and generate the dataset was developed, called Graphlaxy, which was used in this work to generate the synthetic training set. Graphlaxy is open source and available at \url{https://github.com/BNN-UPC/graphlaxy}.

\subsection{Use Case Description}
The scenario that we use to demonstrate the use and results of this method consists of comparing two designs that do not affect the GNN accuracy but that do affect the training time. The designs consist of two different ways of processing the GNN, based on two graph representations: SPARSE and EDGE\_LIST. %More details on the difference between these two representations were explained on Section \ref{sec:preGNN}. 
%We will compare in this study two ways of representing the graph, that change the way the GNN is processed. %One is the sparse adjacency matrix representation (SPARSE), and the other one is the edge list representation (EDGE\_LIST). 
The SPARSE representation uses sparse matrix multiplication to compute the aggregation function, whereas the EDGE\_LIST representation uses \textit{gather} and \textit{scatter} CUDA instructions to compute the aggregation function.

 \begin{figure*}
    \includegraphics[width=\columnwidth]{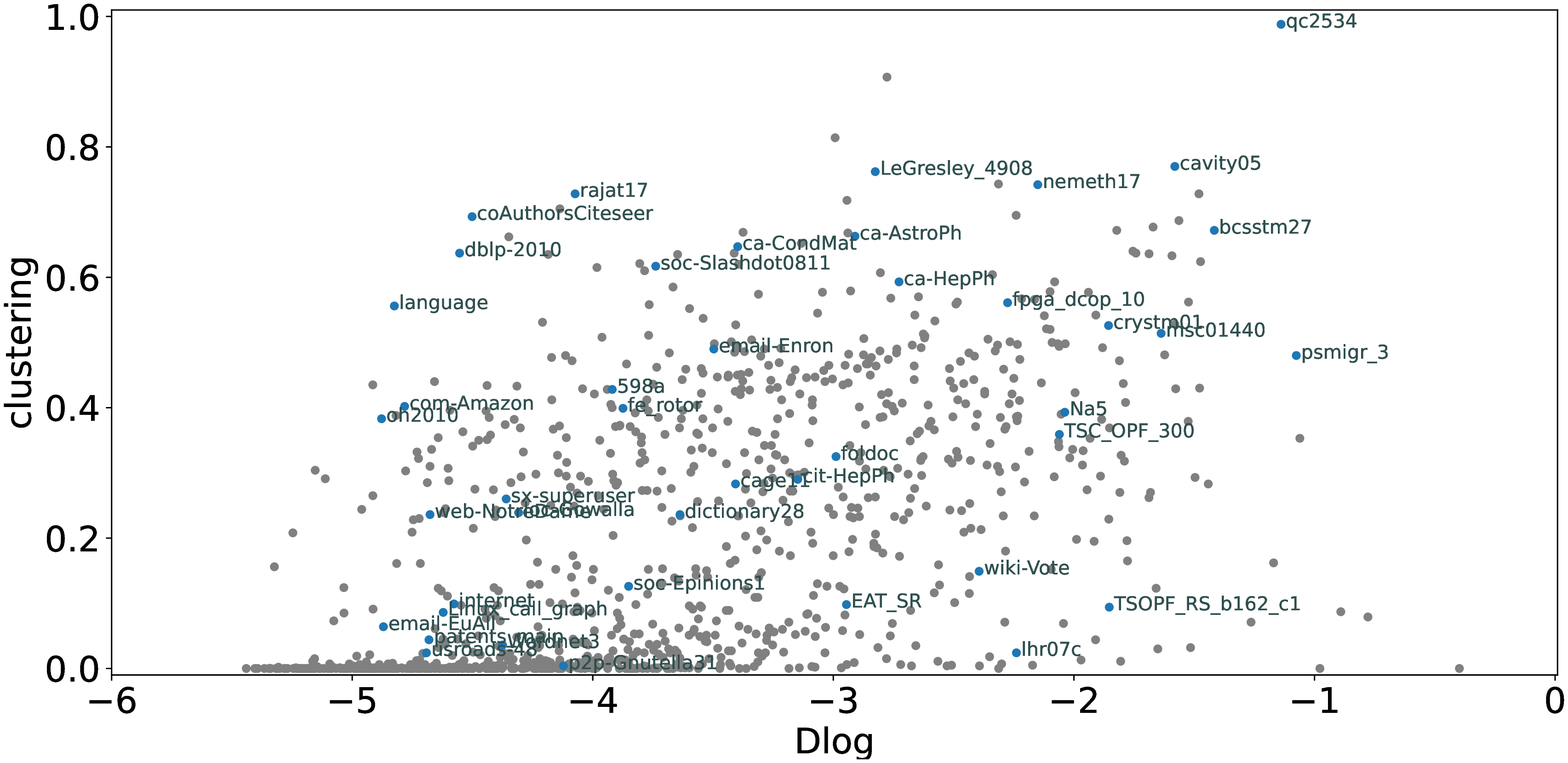}
    \includegraphics[width=\columnwidth]{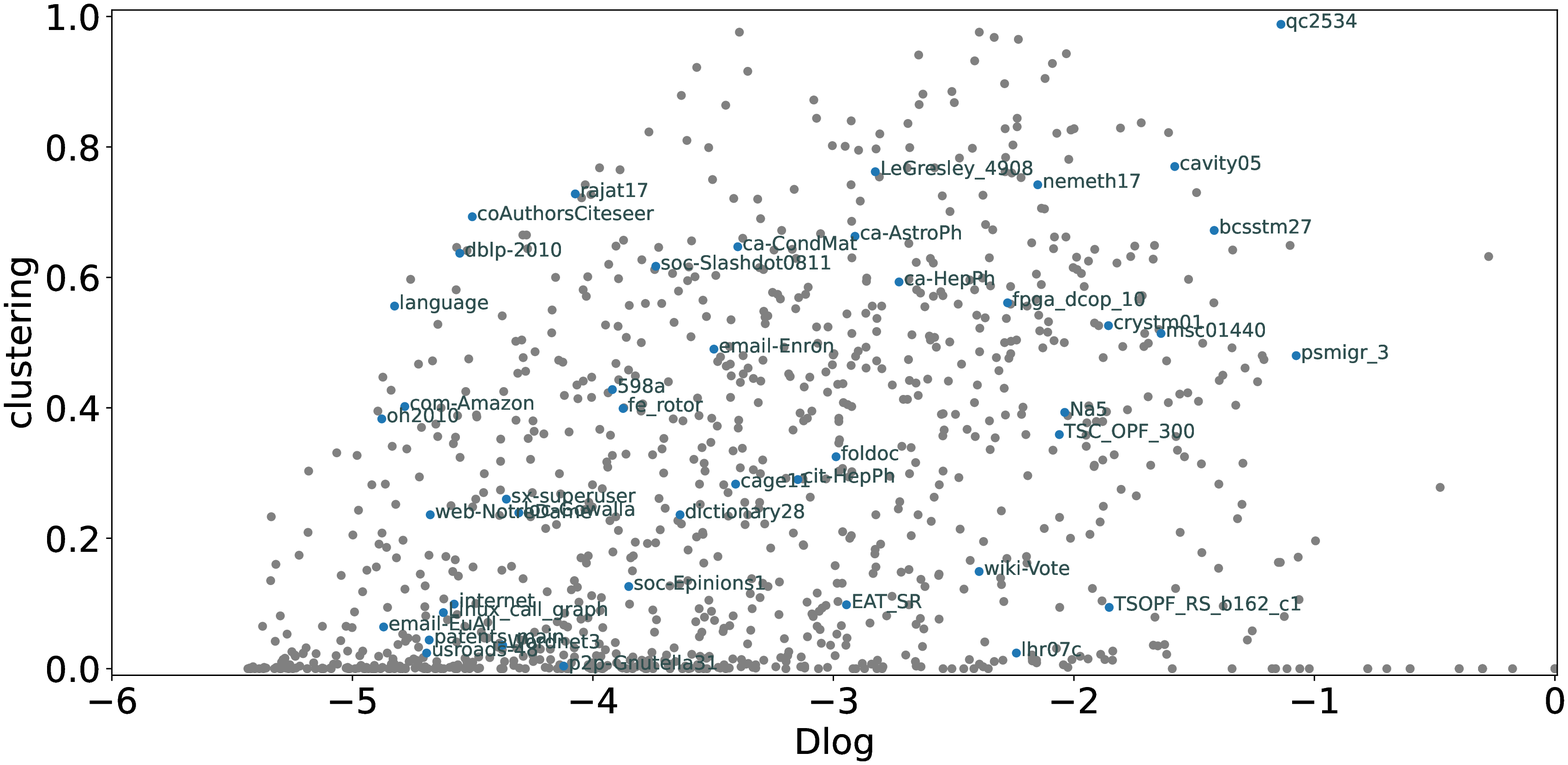}
    \vspace{-0.2cm}
    \caption{Assesed metrics of the synthetic graph datasets (gray), and validation set of real graphs (blue). To the left the naive approach, and to the right the dataset optimized to reduce selection bias. Both are a random sample of the entire ddataset composed of 1000 graphs each to better show the bias.}
    \label{fig:rmat}
\end{figure*}

The experiment was repeated for the four different GNN models, described in Section \ref{sec:preGNN}, to show how the results vary depending on the model. To do a fair comparison, all graphs (training and testing set) were populated with a randomly generated set of features and a random class. The feature vectors are assumed of size 32, the hidden layer of size 32, the number of layers is 3 and the number of classes is 2. 

All the experiments are run using the same \textbf{software}, the framework (PyG) with CUDA version 10.1 and torch version 1.10.2; as well as the same \textbf{hardware}, a machine with CPU Intel(R) Core(TM) i7-2600 CPU @ 3.40GHz, GPU GeForce GTX 980 Ti and 15 GB of RAM. %Also all experiments were run on the GPU.
To reason about the results obtained, executions were profiled using NVIDIA Nsight.

\section{Performance Evaluation}
\label{sec:results}
In this section, we go through the different steps of the methodology showing the results obtained for the training and testing sets and a short discussion about the intuition behind those results. 

\subsection{Dataset Generation}
Using the method described in Section \ref{sec:gen}, we were able to generate a dataset that represents most of the real graphs used as validation. It can be seen in Figure \ref{fig:rmat} that, in the naive dataset created with RMAT only, most of the graphs have low clustering coefficients and that some of the real graphs from the validation set have high clustering coefficients, rendering these graphs underrepresented. We can also see that the final dataset generated with Graphlaxy \cite{Wassington2022BiasRV} solves this problem and that most graphs from the validation set fall inside the limits of the point cloud corresponding to the graphs in this dataset. The dataset is composed of graphs with edges ranging from one thousand to one million. The number of nodes and the rest of the parameters of RMAT are controlled by a distribution resulting from the optimization problem. Such a broad and balanced dataset will allow us to train a regression that takes into account the relationship between the different variables considered.  

\subsection{Analysis}
Once we have an unbiased dataset, an analysis of the impact of the graph metrics on the computation time can be made.
%For this work, we have studied the impact of different metrics on the time it takes for a GNN to process one training generation, and we have come up with a subset of metrics that we found to have the largest impact on the performance.
Since the SPARSE representation of the graph is based on sparse matrix multiplication, we pulled some ideas on which are the most impactful metrics in studies on matrix multiplication \cite{castano2008performance, yeom2016data}. In this work, the most impactful metric was the number of non-zero elements, which is translated to graphs as the number of edges. Additionally, through the experiments, we found that the most relevant metrics were the number of nodes, the number of edges, the maximum degree, and the clustering of the graph. 

\begin{figure}[b]
    \centering
    \includegraphics[width=\columnwidth]{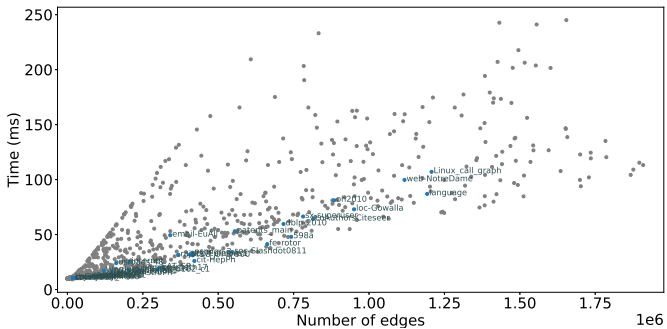}
    \vspace{-0.4cm}
    \caption{Correlation between number of edges and computation time for the GCN model using edge list graph representation for the synthetic training dataset (gray) and the validation dataset (blue).}
    \label{fig:edge_time}
\end{figure}

In Figure \ref{fig:edge_time}, we can see how the number of edges impacts the computation time of the training set and the validation set. We can see also that there is no linearity when the graphs are small. Also, we see that the variation increases the more edges the graph has, indicating that other metrics could explain that variation. A similar figure can be plotted for all the analyzed configurations, resulting in a similar shape. The impact of these metrics on each of the models and graph representations varies, effectively generating a breakpoint between when each of the designs is preferred, as we see later.

The impact of the number of nodes and number of edges can be explained because they indicate the number of operations on the aggregation, though their influence is different for SPARSE and EDGE\_LIST representations. The impact of the maximum degree lies in the fact that highly connected nodes become a bottleneck where all computations of the aggregation on the neighbors must be completed before the calculation of the update. Also, the clustering may be an indicator of the complex dependency between the calculations of the nodes. The non-linearity shown at the left of the graph may be explained by the fact that small graphs use a small portion of the GPU memory and pipeline, and thus slightly bigger graphs use more GPU resources over the same amount of time instead of the same amount of resources during more time. This non-linearity will vary with the hardware used.

\subsection{Regression}

% Please add the following required packages to your document preamble:
% \usepackage{multirow}
\begin{table}
\caption{Scores for the regression for the different configurations considered on this work. \label{tab:reg}}
\centering
\makebox[0.8\columnwidth]{
\begin{tabular}{ll|rrr|rrr}
\multicolumn{2}{l|}{\textbf{Configuration}}     & \multicolumn{3}{l|}{\textbf{Training}}                                                                                    & \multicolumn{3}{l}{\textbf{Testing}}                                                                                  \\
\textbf{Model}        & \textbf{Repr.} & \multicolumn{1}{l}{\textbf{R\textasciicircum{}2}} & \multicolumn{1}{l}{\textbf{MSE}} & \multicolumn{1}{l|}{\textbf{MAPE}} & \multicolumn{1}{l}{\textbf{R\textasciicircum{}2}} & \multicolumn{1}{l}{\textbf{MSE}} & \multicolumn{1}{l}{\textbf{MAPE}} \\ \hline
\multirow{2}{*}{GCN}     & Sparse           & 0.99                                              & 3                                & 0.03                               & 0.98                                              & 11                               & 0.04                              \\
                         & Edge List        & 0.98                                              & 37                               & 0.06                               & 0.97                                              & 21                               & 0.09                              \\ \hline
\multirow{2}{*}{GIN}     & Sparse           & 0.99                                              & 2                                & 0.04                               & 0.99                                              & 1                                & 0.05                              \\
                         & Edge List        & 0.97                                              & 68                               & 0.06                               & 0.98                                              & 19                               & 0.09                              \\ \hline
\multirow{2}{*}{GAT}     & Sparse           & 0.99                                              & 3                                & 0.02                               & 0.99                                              & 18                               & 0.03                              \\
                         & Edge List        & 0.99                                              & 57                               & 0.04                               & 0.97                                              & 50                               & 0.07                              \\ \hline
\multirow{2}{*}{SAGE}    & Sparse           & 0.99                                              & 2                                & 0.06                               & 0.99                                              & 1                                & 0.07                              \\
                         & Edge List        & 0.97                                              & 72                               & 0.08                               & 0.97                                              & 20                               & 0.11                             
\end{tabular}
}
\end{table}

\begin{figure*}
    \centering
    \includegraphics[width=\columnwidth]{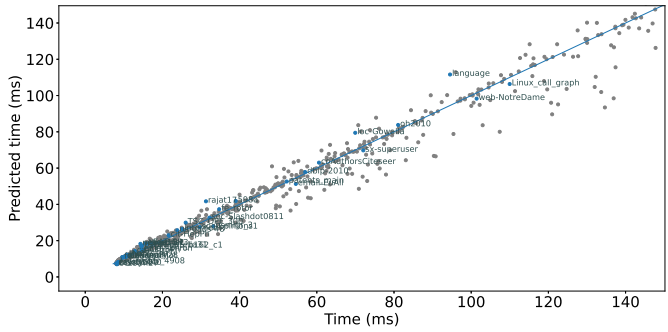}
    \includegraphics[width=\columnwidth]{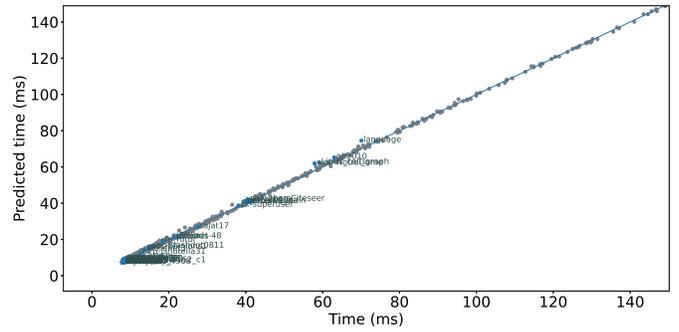}
    \vspace{-0.2cm}
    \caption{Predicted against real time for both graph representations (to the left edge list and to the right saprse representation) in the GIN model case for training dataset (gray) and testing dataset (blue).}
    \label{fig:reg}
\end{figure*}

\begin{figure}
    \centering
    \includegraphics[width=\columnwidth]{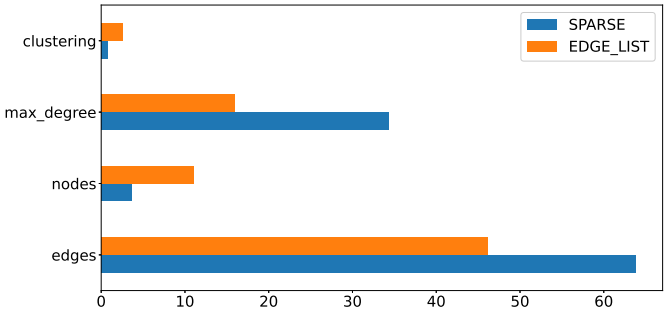}
    \vspace{-0.5cm}
    \caption{Mean impact factor of the different metrics on the regression.}
    \label{fig:impf}
\end{figure}

Based on the analysis we made, we built a model using the metrics (number of nodes, number of edges and maximum degree, and the mean degree) able to predict the training time of the different designs with high accuracy (mean $R^2$ of 0.98 in both training and testing set). The model used is a mix of linear regression with ridge normalization and SVM regression with radial basis function kernel (RBF). The SVM is used to predict the residuals of the linear regression to account for the non-linearity. This compound approach was taken because the SVM by itself is unable to generalize to bigger graphs, and the linear regression is unable to fit the non-linear parts of the data.

%The model used is a linear regression because it was the model that better extrapolates to graphs that are bigger than the ones on the training set. Other models that were evaluated are SVM with different kernels as radial basis function kernel (RBF) and polynomial, and a different set of parameters, which yielded better accuracy but are unable to generalize. 

Table \ref{tab:reg} shows that the results for SPARSE graph representation are in general better than the ones with EDGE\_LIST. We see that the MSE is the value that increases more on the validation set for EDGE\_LIST indicating that some values have higher errors. In general, we can conclude that the models work well for both representations, though slightly better for SPARSE representation. This can be explained in Figure \ref{fig:reg}, where we see that the EDGE\_LIST representation has a higher variation for the same set of metrics, maybe indicating that one more metric could be used, or because of the implications of the method like the order in which the nodes are processed may impact the performance.

To corroborate what we have observed in the analysis, Figure \ref{fig:impf} shows the impact factor, defined as the standard deviation of the variable times the coefficient of the linear regression for each metric. We can see that the number of edges is the more impactful metric, followed by the maximum degree and that the impact of the metric varies from one representation to the other.

\subsection{Classification}

\begin{table}
\centering
\caption{Accuracy of classification of the fastest among two graph representations and speedup obtained with respect to randomly selecting a graph representation.\label{tab:class}}
\begin{tabular}{l|rr|rr}
\multirow{2}{*}{\textbf{Model}} & \multicolumn{2}{l|}{\textbf{training}}                                        & \multicolumn{2}{l}{\textbf{testing}}                                         \\
                                & \multicolumn{1}{l}{\textbf{accuracy}} & \multicolumn{1}{l|}{\textbf{speedup}} & \multicolumn{1}{l}{\textbf{accuracy}} & \multicolumn{1}{l}{\textbf{speedup}} \\ \hline
GCN                             & 0.96                                  & 1.24                                  & 0.90                                  & 1.07                                 \\
GIN                             & 0.93                                  & 1.30                                  & 0.95                                  & 1.26                                 \\
GAT                             & 0.97                                  & 1.16                                  & 0.90                                  & 1.05                                 \\
SAGE                            & 0.95                                  & 1.35                                  & 0.98                                  & 1.35                   
\end{tabular}
\end{table}

Once the regression models are built for each design, the results of the models can be used to classify the graphs by which design will perform better.

\begin{figure}
    \centering
    \includegraphics[width=\columnwidth]{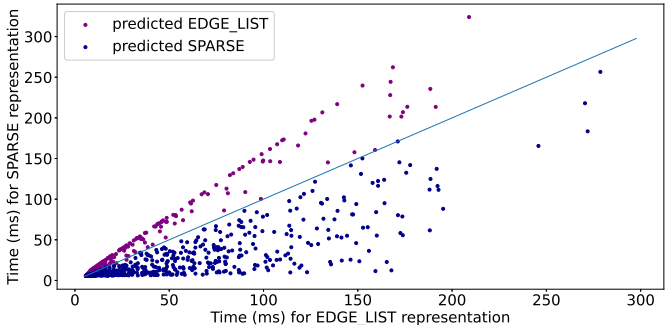}
    \vspace{-0.5cm}
    \caption{Scatter plot with the training time, per epoch, for each of the graph representations (sparse and edge list) for the GraphSAGE model. The diagonal line in blue indicates the frontier where both representations lead to the same processing time, whereas the color of the dots represent the prediction made by \textsc{ProGNNosis}.}
    \label{fig:class}
\end{figure}

From Table \ref{tab:class}, we can obtain the accuracy of the classification between graphs that will work better with the SPARSE representations and the ones that will work better with the EDGE\_LIST representations. These results are good all over 0.9. From Figure \ref{fig:class}, we can see that the graphs that are not being correctly classified are close to the diagonal, meaning that the time for both graph representations is similar. Therefore, classifying them wrongly may have a low impact on the GNN acceleration results.

\subsection{Use Case Summary}

\begin{figure}
    \centering
    \includegraphics[width=\columnwidth]{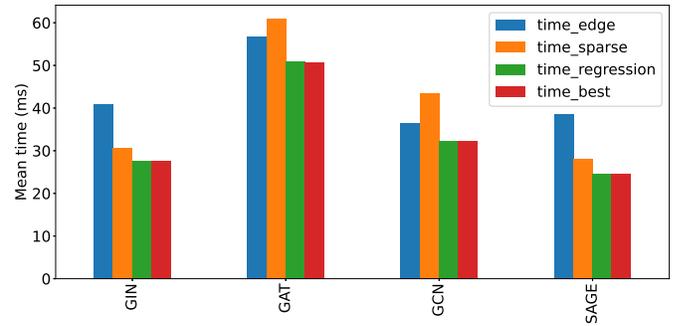}
    \vspace{-0.5cm}
    \caption{Mean training time for multiple strategies. \emph{time\_edge} refers to always using the edge list representation, \emph{time\_sparse} refers to always using the sparse matrix representation, \emph{time\_regression} plots the time obtained with the regressions of \textsc{ProGNNosis} and, finally, \emph{time\_best} represents the ideal case that always selects the fastest option.}
    \label{fig:time_bar}
\end{figure}

Using the designed method, we demonstrate through Figure~\ref{fig:time_bar} that GNN computation can be accelerated through informed decisions. By using \textsc{ProGNNosis}, we can achieve a speedup close to the one we would obtain by always selecting the best option between the two designs. A summary of the obtained speedups is also shown in Table \ref{tab:class}. Models such as GAT or SAGE showed potential speedups over 1.30$\times$. We have also seen that when sweeping other variables out of the scope of this study, such as the number of input features, the tradeoffs vary but, still, \textsc{ProGNNosis} allows to identify the best representation strategy. However, the results are not shown in the paper for the sake of brevity.

\section{Conclusion}
\label{sec:conc}
In this study, we have analyzes the impact of different graph metrics on the training time of different GNN models and different graph representations. With these results, we were able to build a model that can predict the training time of different GNN designs. Furthermore, we demonstrated that this method can be used to select the best design in terms of training time for a given GNN model and a graph of arbitrary characteristics. In the presented scenario, we are only comparing the processing time assuming two graph representation alternatives and, even so, we achieved significant speedups of over 1.20$\times$ on average and exceeding 1.30$\times$ in specific models. However, with a wider design space, the speedups could be increased. In future work, we intend to include the feature vector size and other variables as part of the regression models. Further, the method can be generalized not only to multiple GNN models, as we have seen in this paper, but also to other design variables such as the internal dataflow of a hardware accelerator \cite{garg2022understanding}.

It is important to notice that the speedups obtained do not take into consideration the time taken to calculate the metrics, the regression, or the time to transform the graph from one representation to another. Yet still, such pre-processing steps only need to be performed once for each graph and may be done offline depending on the scenario.  Also, a tradeoff may be optimized between the approximation used for calculating the metric and the accuracy of the regression. These are left as future work.

\ifCLASSOPTIONcaptionsoff
  \newpage
\fi

% trigger a \newpage just before the given reference
% number - used to balance the columns on the last page
% adjust value as needed - may need to be readjusted if
% the document is modified later
%\IEEEtriggeratref{8}
% The "triggered" command can be changed if desired:
%\IEEEtriggercmd{\enlargethispage{-5in}}

% references section

% can use a bibliography generated by BibTeX as a .bbl file
% BibTeX documentation can be easily obtained at:
% http://mirror.ctan.org/biblio/bibtex/contrib/doc/
% The IEEEtran BibTeX style support page is at:
% http://www.michaelshell.org/tex/ieeetran/bibtex/
\bibliographystyle{IEEEtran}
% argument is your BibTeX string definitions and bibliography database(s)
\bibliography{IEEEabrv,main}
%
% <OR> manually copy in the resultant .bbl file
% set second argument of \begin to the number of references
% (used to reserve space for the reference number labels box)
%\begin{thebibliography}{1}

%\bibitem{IEEEhowto:kopka}
%H.~Kopka and P.~W. Daly, \emph{A Guide to {\LaTeX}}, 3rd~ed.\hskip 1em plus
%  0.5em minus 0.4em\relax Harlow, England: Addison-Wesley, 1999.

%\end{thebibliography}

% biography section
% 
% If you have an EPS/PDF photo (graphicx package needed) extra braces are
% needed around the contents of the optional argument to biography to prevent
% the LaTeX parser from getting confused when it sees the complicated
% \includegraphics command within an optional argument. (You could create
% your own custom macro containing the \includegraphics command to make things
% simpler here.)
%\begin{IEEEbiography}[{\includegraphics[width=1in,height=1.25in,clip,keepaspectratio]{mshell}}]{Michael Shell}
% or if you just want to reserve a space for a photo:

% \begin{IEEEbiography}{Michael Shell}
% Biography text here.
% \end{IEEEbiography}

% if you will not have a photo at all:
\begin{IEEEbiographynophoto}{Axel Wassington}
\hl{Biography text here.}
\end{IEEEbiographynophoto}

% insert where needed to balance the two columns on the last page with
% biographies
%\newpage

\begin{IEEEbiographynophoto}{Sergi Abadal}
\hl{Biography text here.}
\end{IEEEbiographynophoto}

% You can push biographies down or up by placing
% a \vfill before or after them. The appropriate
% use of \vfill depends on what kind of text is
% on the last page and whether or not the columns
% are being equalized.

%\vfill

% Can be used to pull up biographies so that the bottom of the last one
% is flush with the other column.
%\enlargethispage{-5in}

% that's all folks
\end{document}